\documentclass[conference]{IEEEtran}
\IEEEoverridecommandlockouts

\usepackage{amsmath,graphicx,hyperref}
\usepackage{cite}
\usepackage{amsmath,amssymb,amsfonts}
\usepackage{algorithmic}
\usepackage{graphicx}
\usepackage{tabularx}
\usepackage{textcomp}
\usepackage{xcolor}
\usepackage{xspace}
\usepackage{booktabs}
\usepackage{svg}
\usepackage{microtype}
\usepackage{url}

\newcolumntype{C}{>{\centering\arraybackslash}X}
\newcolumntype{L}{>{\raggedright\arraybackslash}X}
\newcolumntype{R}{>{\raggedleft\arraybackslash}X}

\def\BibTeX{{\rm B\kern-.05em{\sc i\kern-.025em b}\kern-.08em
    T\kern-.1667em\lower.7ex\hbox{E}\kern-.125emX}}

\usepackage{pifont} %
\newcommand{\vmark}{\ding{51}}
\newcommand{\xmark}{\textcolor{gray!25}{\ding{55}}}

\newcommand\etal{\textit{et al.}\xspace}
\newcommand\eg{\textit{e.g.}\xspace}
\newcommand{\ab}{\mathbf{a}}
\newcommand{\ib}{\mathbf{i}}

\usepackage[prependcaption,textsize=scriptsize]{todonotes}
\definecolor{mycolor}{HTML}{FF6600}
\definecolor{indiagreen}{HTML}{138808}
\definecolor{papaya}{HTML}{EE892F}
\definecolor{mygreen}{HTML}{008000}
\definecolor{mypurple}{HTML}{9966CC}
\definecolor{myblue}{HTML}{5D8AA8}
\definecolor{mypink}{HTML}{EC008C}

\newcommand{\herman}[1]{\textcolor{black}{#1}}

\newcommand{\dan}[1]{\textcolor{black}{#1}}

\begin{document}

\title{Connecting Speech to Words through Images
\thanks{%
This work was supported in part by two grants of the Ministry of Research, Innovation and Digitization, CNCS-UEFISCDI, project numbers PN-IV-P2-2.1-TE-2023-1632 and PN-IV-P7-7.1-PTE-2024-0546, within PNCDI IV.
}
}

\author{
  \IEEEauthorblockN{Gabriel Pirlogeanu, Dan Oneata, Horia Cucu}
  \IEEEauthorblockA{\textit{Speech and Dialogue Research Laboratory} \\
  \textsc{Politehnica} \textit{Bucharest}, Romania\\
  \{gabriel.pirlogeanu, dan\_theodor.oneata, horia.cucu\}@upb.ro}\\
  \and
  \IEEEauthorblockN{Herman Kamper}
  \IEEEauthorblockA{\textit{Electrical and Electronic Engineering} \\
  \textit{Stellenbosch University,} South Africa \\
  kamperh@sun.ac.za}
 }
 
\maketitle

\begin{abstract}
How can we learn the mapping between written words and their spoken counterparts in the absence of explicit textual supervision? We present a visually grounded method for building a vocabulary of spoken words using only images and their spoken descriptions. First, image captioning systems are used to build a vocabulary of written words representing salient visual concepts in the images. For each word, we then find utterances whose image captions contain that word. Then we use an unsupervised word discovery technique to align these utterances to locate instances of the target word. The result is spoken word segments that are linked to written words---all accomplished without any text supervision. In spoken word retrieval and keyword spotting experiments, the proposed approach outperforms a strong neural baseline while being more interpretable. These results demonstrate the feasibility of the approach in English and motivate future work on low-resource languages without transcripts.
\end{abstract}

\section{Introduction}

One way to collect speech resources is through visual grounding: 
start with a set of images, ask people to describe them, and record their utterances. 
This can be done even for languages where literacy rates are low, the language does not have a written form, or when collecting transcriptions is prohibitively expensive~\cite{vaani2025}.

In this work we ask whether it is possible to link audio segments to written words using such a visually grounded dataset.
Previous work has explored using visual grounding to map speech to image regions \cite{harwath2018eccv} or even speech to speech across languages \cite{azuh2019interspeech},
but less emphasis has been placed on mapping speech to written words.
Connecting speech to words establishes a clear semantic correspondence,
and could support language documentation by ascribing meaning to speech segments from a foreign language.

We %
propose using off-the-shelf image captioning systems to obtain textual supervision from images.
For each image in the dataset, we generate a written description using several automatic image captioning systems.
\herman{From this,} we can generate a vocabulary of written words that has a visual correspondence.
To map a %
keyword to an audio segment, we take inspiration from unsupervised word discovery~\cite{vanniekerk24_interspeech}.
The idea is to run a speech segmentation method only on those utterances that are likely to contain a given word:
\herman{utterances whose corresponding captions contain the target word are selected, and each pair of utterances is aligned using a non-parametric approach.}
The segments that agree most with the other utterances are deemed to contain the word of interest.
For each vocabulary word, we obtain speech segment spans predicted to contain that keyword.

Olaleye \etal \cite{olaleye2022jstsp} proposed an approach for the related task
of keyword localization in visually grounded speech:
given an image--speech corpus and a written keyword, locate the keyword in the speech corpus.
Their main idea was to use an image tagger to provide weak supervision for an audio-to-keyword network.
The keywords that can be retrieved are limited and fixed to the codomain of the single image tagger used.
In our case, the vocabulary is dynamically created, based on the corpus at hand.

\begin{figure}[t]
    \centering
    \includegraphics[width=0.85\linewidth]{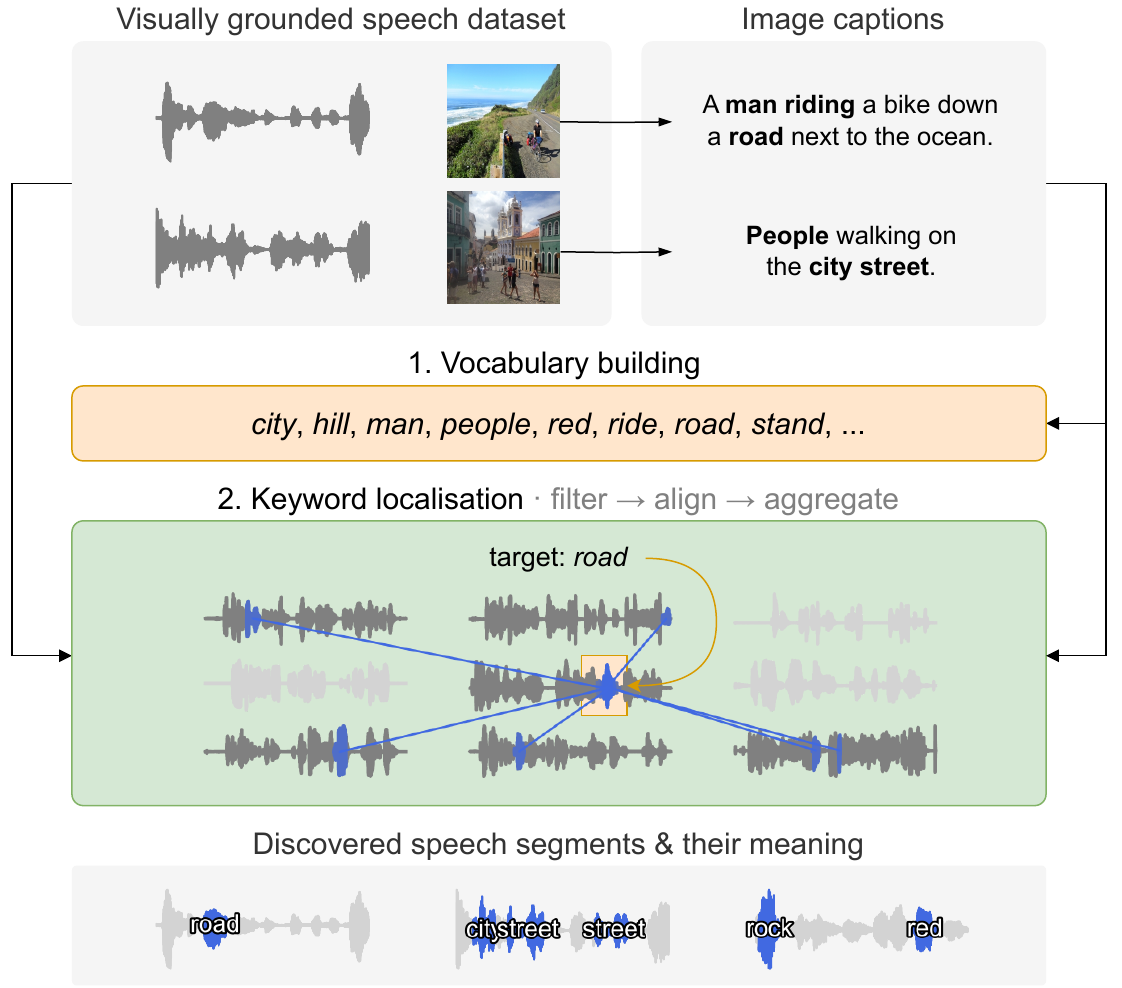}
    \caption{
        Given %
        a visually grounded speech corpus, we 
        learn a mapping between written words and their spoken counterparts.
        We construct a vocabulary from the most frequent words from %
        captions, obtained automatically for the images.
        We find all utterances whose corresponding image captions contain a target word, %
        align these utterances and aggregate the results to
        retrieve the most aligned audio segments. %
        }
    \label{fig:overview}
\end{figure}

We evaluate two variants of our approach: %
one variant of the unsupervised word segmentation system uses continuous self-supervised features, while the other relies on discrete features. We show that they obtain similar results, both outperforming
an updated version of the parametric method of Olaleye \etal \cite{olaleye2022jstsp}.
Through a series of ablation studies and analyses, we identify the strengths and shortcomings of our approach.
Since the generated captions are not perfectly aligned with what is being said, %
we find that it is important to use multiple captioning systems.
We also identify particular failure modes of our approach.

This study lays the foundation for our follow-up work: on datasets~\cite{vaani2025,olaleye2022yfaccyorubaspeechimagedataset} where the spoken captions are in a low-resource language. We aim to learn associations between foreign speech and English words, supporting language documentation efforts in scenarios where textual transcriptions cannot be collected.

\section{Methodology}
\label{sec:methodology}

We assume we have %
a visually grounded speech dataset,
where images are paired with corresponding spoken descriptions.
Using this data,
our goal is to discover acoustic segments and assign them meaning in the form of written words.

Our approach, illustrated in Fig.~\ref{fig:overview}, consists of two main steps. %
We first
define a vocabulary to represent the target meaning space. %
Then we identify the corresponding audio segments for each word in the vocabulary using a novel keyword localization method.  
Through this process, each vocabulary word becomes associated with a set of relevant audio segments---effectively linking spoken language to written words.  
This %
association is achieved using only visual context for supervision, without requiring %
any textual
annotations.

For the vocabulary building step, we use pretrained image captioning models to generate descriptions for the available images.
We lemmatize the captions and assemble %
a list of the most frequent words.

For the keyword localization step, we assume that we are given a written query word and we want to retrieve relevant audio segments. %
We propose a new keyword localization method that takes inspiration from unsupervised word discovery~\cite{interval_piling,R2018254,dunbar2021zero,vanniekerk24_interspeech},
but constrains the search space to audio files linked to the query word.
Specifically, we filter %
utterances based on their associated image captions, retaining only those that contain the query (Sec.~\ref{sec:filter}).
Then, we align the selected utterances with one another (Sec.~\ref{sec:align}) and
identify the recurring segments using an interval piling technique (Sec.~\ref{sec:rank}).
The recurring segments are assumed to contain the query word.

\subsection{Filtering by Visual Information}
\label{sec:filter}

We first select audio--image pairs that are likely to contain the written query word $w$.
This %
is %
based on the visual information using the automatically generated image captions: if the image caption contains %
$w$, then the audio--image pair $(\ab, \ib)$ is retained: \(\left\{ (\ab, \ib) \; | \; w \in \mathrm{ImageCaptioner}(\ib) \right\}\).
We use the same image captioning systems as for the vocabulary building step, and we also lemmatize the words in the captions.

\subsection{Aligning Utterances}
\label{sec:align}

\begin{figure}[t]
    \centering
    \scalebox{0.5}{
	    \includegraphics{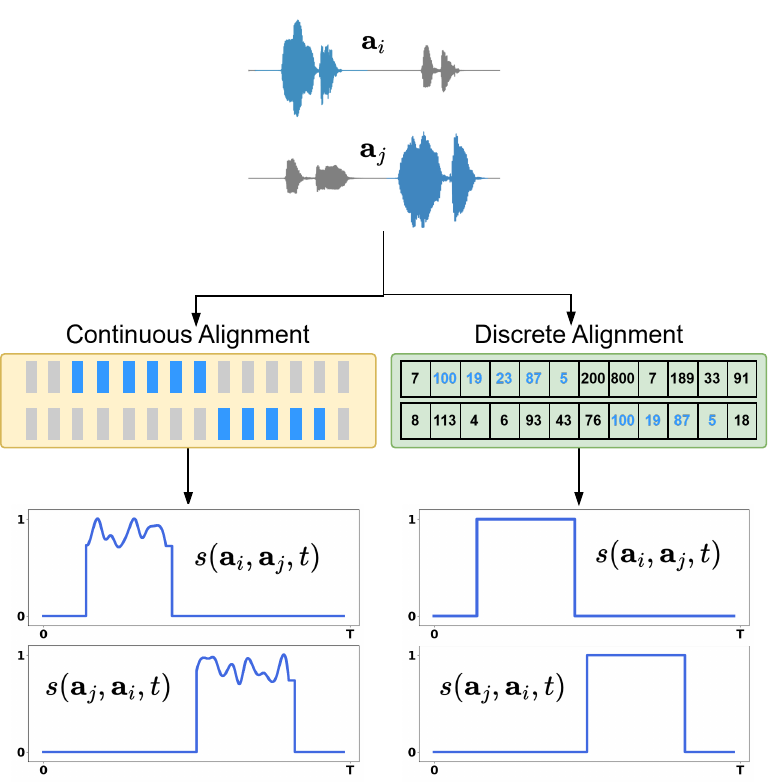}}%svg}}
    \caption{
        Continuous and discrete alignment methods are applied to a pair of input utterances.
        The discrete method takes cluster codes as input and produces a binary alignment signal. The continuous approach yields a continuous similarity score in the range \([0, 1]\).
    }
    \label{fig:comparison_methods}
\end{figure}

Next, the audio files selected for a particular word are aligned to each other.
The idea is that this target word would appear in all the utterances, and thus correspond to the most frequently occurring common audio subsequence.
To extract these subsequences, we take inspiration from the unsupervised word discovery approach of~\cite{azuh2019interspeech}. We explore two variants: one based on continuous self-supervised representations, and the other on discrete ones.

For
any two audio utterances $\ab_i$ and $\ab_j$, we define a scoring function $s(\ab_i, \ab_j, t)$ for %
how likely it is that segment $t$ from %
utterance $\ab_i$ appears anywhere in %
utterance $\ab_j$.
We compute the score based on either continuous features or discrete features, as illustrated in Fig.~\ref{fig:comparison_methods}.
Both %
variants rely internally on HuBERT representations \cite{10.1109/TASLP.2021.3122291}. %

\textit{Discrete features alignment (DFA):}
As in \cite{vanniekerk24_interspeech},
HuBERT features are first encoded to discrete units using k-means clustering, and then each pair of unit sequences are
aligned using the Smith-Waterman dynamic programming algorithm~\cite{smith-waterman}.
From the resulting alignment, we construct a binary scoring function $s$:
1 if the segment in $\ab_i$ is matched, and 0 otherwise.

\textit{Continuous features alignment (CFA):}
Following \cite{azuh2019interspeech}, we estimate the alignment of two utterances by computing the similarity directly in the feature space.
We use cosine similarity between features extracted at various time frames,
and define the score between a pair of audios as the maximum similarity between the features $\phi_{it}$, extracted from the $\ab_i$ at time $t$, and all the other features $\phi_j$ from the second audio: \(s(\ab_i, \ab_j, t) = \max_{t'} \langle \phi_{it}, \phi_{jt'} \rangle\).
To obtain a less noisy and sparse signal,
we apply a smoothing filter (using a standard Gaussian) and set the values below \( \gamma \cdot \max (s) \) to 0, where $\gamma$ is a hyperparameter. %
While continuous features have more capacity than discrete ones, framewise comparisons are much slower.

\subsection{Ranking Segments by Aggregation}
\label{sec:rank}

Given the alignments between %
the audio files linked to a query word, we want to locate the word in those utterances.
The idea is that a segment within a particular utterance that is matched to several other utterances in the %
set
is likely to contain the word.
So, we define an utterance-wise score that aggregates the alignment scores between each audio $\ab_i$ and all other audio utterances $\ab_j$: \(s(\ab_i, t) =  \sum_{j \neq i} s(\ab_i, \ab_j, t)\).
This is called %
``interval piling''~\cite{interval_piling}.
A higher value in the aggregated alignment \( s(\ab_i, t) \) indicates that the corresponding region in utterance \( \ab_i \) aligns more frequently with segments from other utterances and it is likely to contain the query word.

To eliminate artefacts resulting from noisy alignments, %
we discard values below %
a threshold $\theta$, which is a fraction of the maximum aggregated score.
The speech is then segmented by using the contiguous regions bounded by silence or zero-valued frames. Each resulting segment is scored based on the average of its frame-level values.

\section{Experimental Setup}
\label{sec:experimental-setup}

\subsection{Data and Vocabulary}
\label{sec:vgs-data}

The MIT Places Audio Captions dataset~\cite{harwath2016nips,harwath2017acl,harwath2018eccv} is a speech--image dataset, of 400k pairs of images with spoken English descriptions.
The images~\cite{NIPS2014_19ea3982} \dan{contain} scenes such as forests, kindergarten classrooms, or car interiors.
The spoken captions were collected via Amazon Mechanical Turk and
are %
spontaneously spoken rather than read. %
On average, a caption has a duration of about 10 seconds with roughly 20 words. For our experiments, we sample 20k speech--image pairs, and use 10k of those samples for development purposes and the other %
10k for %
final evaluation.
The original dataset doesn't provide manual transcriptions.
To perform our evaluation (Sec.~\ref{sec:eval}), we need accurate transcriptions as well as forced alignments.
We obtain transcriptions using \dan{a}
state-of-the-art ASR system, the Parakeet TDT-CTC 110M model \cite{10389701} from NVIDIA’s NeMo toolkit. %
We align the transcriptions using a CTC-based forced aligner %
using the Massively Multilingual Speech (MMS) model~\cite{10.5555/3722577.3722674}.

We build a vocabulary from words that are likely present in the images from our corpus.
We describe the images using three state-of-the-art image captioning systems:
Tag2Text~\cite{huang2023tag2text}, BLIP-2~\cite{10.5555/3618408.3619222}, and GIT~\cite{wang2022gitgenerativeimagetotexttransformer}.
For each system we generate a caption using beam search.
Stopwords are removed and words lemmatized using the \texttt{en\_core\_web\_sm} SpaCy model~\cite{Honnibal_spaCy_Industrial-strength_Natural_2020}.
The final set of words associated with each image is the intersection of the words produced by all three captioners for that image.
The vocabulary contains the most frequent 100 words,
making sure we don't include visually ungrounded terms such as ``background'' or ``picture''.
This process is run independently on both development and test splits.

\subsection{Implementation Details}
\label{sec:implementation-details}

For both alignment methods, we extract features from the seventh layer %
of the English HuBERT model, the optimal layer for phone discrimination~\cite{10.1109/TASLP.2021.3122291}.
To avoid \dan{aligning} silence or background noise \dan{across} utterances,
we apply voice activity detection using Pyannote3~\cite{Plaquet23}.
The alignment %
hyperparameters are tuned on the \texttt{dev} split using the intersection of the three captioning systems. %
For DFA, we set %
the aggregated score threshold $\theta = 0.4$.
For the alignment step, we use same hyperparameters as in \cite{vanniekerk24_interspeech},
except for the similarity threshold $\tau$, which we set to 4;
this threshold specifies whether two substrings match.
For CFA, we set the local alignment threshold $\gamma = 0.7$ and the aggregated score threshold $\theta = 0.3$ .
The DFA method is efficient and can run on the CPU.
CFA on the other hand requires matrix multiplication, so we run it on a GPU (Tesla T4).
Aligning 250 audios with CFA and DFA takes around $2$m$15$s and $24$s, respectively.

\subsection{Topline and Alternative Approach}
\label{sec:baseline}

\hspace{\parindent}
\textit{Transcript topline:}
Image captions only approximate the words %
in an %
utterance.
To understand the system's potential under ideal conditions, we evaluate its performance using the utterance's transcript.

\textit{Attention CNN:}
As an alternative approach, we consider our own updated version of the %
neural-based model of
Olaleye \etal \cite{Olaleye2021AttentionBasedKL}, the only other work to look at a related task.
The model has two inputs, an audio utterance and a word from the vocabulary, and predicts whether that word appears anywhere in the utterance (based on the associated image captions).
We use HuBERT features from the seventh %
transformer layer as input, to ensure a fair comparison with our method.
The model
consists of
a convolutional neural network, an attention layer (to temporarily pool the audio embeddings based on the input word),
and a two-layer perceptron (to project the pooled embedding to a binary prediction).
To localize 
a word, we find the peak of the attention weights and return a fixed-sized segment around the peak (from 0.1s before to 0.3s after the peak).
This is necessary since the attention plots are very peaky, resulting in poor performance for this approach on our benchmark (\cite{Olaleye2021AttentionBasedKL} only evaluated whether the peak occurred within the true word, not whether a segment can be extracted). The displacement hyperparameters
were tuned on the \texttt{dev} set.

\subsection{Evaluation}
\label{sec:eval}

Given a retrieved audio segment for a query word, we consider the prediction correct if the segment overlaps more than a certain threshold with a segment containing the query word.
To measure the overlap, we use the intersection over union (IoU) of the two segments.
As ground truth, we use the ASR transcripts and their forced alignments (Sec.~\ref{sec:vgs-data}).
As a secondary metric, we also report the keyword spotting performance.
A retrieved audio segment is assessed as correct if the corresponding utterance contains the query word anywhere.
This metric is a strict %
upper-bound on the localization performance.
For both metrics, we take the top 10 retrieved audio segments for each word in the vocabulary and report the number of correct predictions (precision at rank 10; P@10). %
The scores are averaged over all words in the vocabulary.

 \section{Experimental Results}
\label{sec:experimental-results}

\subsection{Main Results}
\label{sec:main-results}

\begin{table}[t]
    \centering
    \caption{%
        Precision at 10 (\%) on keyword localization and spotting
        using %
        continuous (CFA) or discrete (DFA) feature alignment,
        and a neural baseline (Attention CNN~\cite{Olaleye2021AttentionBasedKL}).
        Toplines use %
        transcripts.
    }
    \begin{tabularx}{\linewidth}{@{}l@{\ \ }c Ccc@{}}
     \toprule
    & &  &\multicolumn{2}{c}{Localization} \\
     \cmidrule(l){4-5}
     Method                                           & Supervision & Spotting       & IoU = 0.5     & IoU = 0.75    \\
     \midrule                                                        
     \multicolumn{5}{@{}l}{\it Toplines} \\                          
     DFA (ours)                                              & Transcripts & \bf 100.0      & 84.5         & 33.2           \\
     CFA (ours)                                              & Transcripts & \bf 100.0      & \textbf{91.7} & \textbf{53.0} \\
     \midrule                                                        
     \multicolumn{5}{@{}l}{\it Visually grounded systems} \\         
     Attention CNN  & Images      & 53.8           & 44.4          & 25.1          \\
     DFA (ours)                                              & Images      & \textbf{85.3}  & 64.3          & 27.7          \\
     CFA (ours)                                              & Images      & 84.4           & \textbf{67.1} & \textbf{35.4} \\
     \bottomrule
    \end{tabularx}
    \label{tab:main-results}
\end{table}

We compare the discrete (DFA) and continuous (CFA) variants of our %
approach with their corresponding toplines (using transcripts) and a neural baseline. %
Table~\ref{tab:main-results} gives %
the results.
All visually grounded methods use the intersection of all three captioning systems %

\textit{Topline performance:}
We observe that using transcripts yields strong results. %
The localization performance for both methods is over 84\% at an IoU of 0.5.
Keyword spotting with transcripts is by definition perfect.
The fact that the words are not always correctly localized is often due to co-occurrences:
\eg, if ``snow'' and ``mountain'' co-occur frequently, the model may predict the one that it's easier to align.
We also observe that CFA has a higher ceiling than DFA, the former showing stronger results at both IoU thresholds in this idealized setting.
A possible reason for the slightly worse performance of DFA is the clustering step, which invariably removes %
information.

\textit{Visually grounded systems:}
Compared to the topline variants, relying on visual content alone leads to an expected drop in absolute performance---24.6\% for CFA and 20.2\% for DFA at an IoU threshold of 0.5.
Even though we use captions, which provide sensible descriptions for images, there is an inherent variability in how images can be described.
This is not only true for machine-generated captions, but also in humans \cite{wang2016lrec}.
CFA once again proves to be the most effective method, %
delivering a 27.7\% relative improvement over DFA in the stricter segmentation scenario.
Both variants outperform the neural approach, thereby improving on the current state-of-the-art.

\begin{figure*}
    \centering
    \begin{tabular}{@{}cccc@{}}  %
	    {\includegraphics[width=0.23\textwidth]{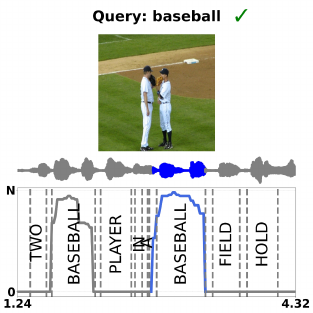}} &%svg}} &
	    {\includegraphics[width=0.23\textwidth]{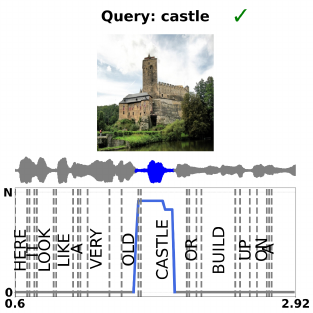}} &%svg}} &
	    {\includegraphics[width=0.23\textwidth]{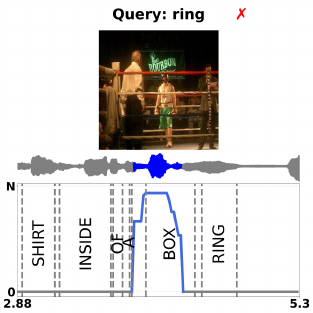}} &%svg}} &
	    {\includegraphics[width=0.23\textwidth]{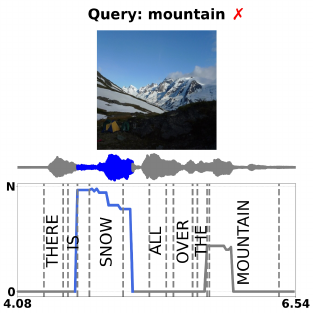}} \\%svg}} \\
    \end{tabular}
    \caption{%
        Examples of retrieved segments for four query words using the DFA method.
    }
    \label{fig:four-images}
\end{figure*}

\textit{Qualitative results:}
Fig.~\ref{fig:four-images} presents output examples from our approach for four query words.
We observe that predictions are generally well aligned with word boundaries, indicating good segmentation. For the word ``baseball'', we see the impact of the ranking algorithm, as it retrieves the segment with the higher average frame-level score.
In the cases where the approach fails, this is related to words that typically co-occur:
\eg, ``box'' gets retrieved for ``ring'', and ``snow'' for ``mountain''. %

\subsection{Ablation Study: Impact of Image Captioning}
\label{sec:cap_perf}

\begin{table}[t]
    \centering
    \caption{%
        Captioning performance (\%), in terms of precision (P) and recall (R), and downstream localization performance, in terms of precision at 10 (at two IoU thresholds).
        We consider all combinations of the three image captioning systems (GIT, BLIP2, Tag2Text) and measure their impact on our two alignment methods (DFA and CFA).
    }
    \begin{tabularx}{\linewidth}{@{}C@{\ \ }CC@{\qquad} cc cc cc@{}}
    \toprule
    & & & \multicolumn{2}{c}{Captioning} & \multicolumn{2}{c}{IoU = 0.5} & \multicolumn{2}{c}{IoU = 0.75}\\
    \cmidrule{4-5} \cmidrule(l){6-7}
    \cmidrule(l){8-9}
    GIT & BLIP2 & Tag2Text & P & R & DFA & CFA & DFA & CFA \\
    \midrule
    \multicolumn{9}{@{}l}{\it Single image captioning system} \\
    \vmark & \xmark & \xmark & 43.5 & \textbf{52.6} & 60.9 &  39.1 & 24.3 & 18.6\\
    \xmark & \vmark & \xmark & 50.1 & 47.1 & 64.0 & 51.0 & 26.7 & 26.6\\
    \xmark & \xmark & \vmark & 49.2 & 46.0 & 61.4 & 52.0 & 26.8 & 25.1\\
    \midrule
    \multicolumn{9}{@{}l}{\it Intersection of two image captioning systems} \\
    \vmark & \vmark & \xmark & 56.3 & 37.8 & \textbf{64.7} & 61.9 & 25.8 & 32.4\\
    \vmark & \xmark & \vmark & 55.8 & 37.0 & 63.2 & 61.4 & 26.1 & 31.1\\
    \xmark & \vmark & \vmark & 57.8 & 36.5 & 63.2 & 63.1 & 26.7 & 33.3\\
    \midrule
    \multicolumn{9}{@{}l}{\it Intersection of three image captioning systems} \\
    \vmark & \vmark & \vmark & \textbf{60.8} & 31.5 & 64.3 & \textbf{67.1} & \textbf{27.7} & \textbf{35.4} \\
    \bottomrule
    \end{tabularx}
    \label{tab:image-captioning-herman}
\end{table}

We observed a performance gap when going from transcripts to visual grounding.
Since the visual information is represented as captions,
we examine how the captioning step impacts final performance.
To assess caption quality, we measure how well a caption matches its corresponding audio transcript.
We report precision and recall: a caption word is considered positive if it also appears in the transcript.
The metrics are averaged over the 100 words in our vocabulary.
Table~\ref{tab:image-captioning-herman} shows the captioning and final localization performance. %

We see that the captioning performance of the three systems is similar, with figures between 40\% and 50\%.
However, the precision--recall trade-off differs: BLIP2 and Tag2Text have the best precision, while GIT has the best recall.
This trade-off impacts the downstream performance: under all four settings,
the two systems that have better precision
localize better than the least precise one. %

This indicates that it is more important to have fewer, but correctly selected utterances than more, but noisier samples.
To further improve the captioning precision, we consider the intersection of more captioning systems:
we select a sample if a query word appears in the selected captioning systems.
As expected, the precision improves (at the expense of recall), with the best option to integrate all three systems.
This gives the best results on localization in three of the four settings.

Between %
discrete and continuous alignment, %
we observe that the discrete DFA has a weaker dependence on the captioning precision,
which is especially noticeable at the lower IoU threshold.
This results in DFA consistently outperforming CFA in the single-captioning setting.
\herman{DFA comes with the additional benefit of being much faster, so realizing its full potential will be the focus of future work.}

\section{Conclusions}

\herman{We proposed}
a non-parametric visually grounded approach for connecting written words to speech segments, and then applied the approach to spoken keyword retrieval.
Our approach extends an unsupervised word discovery method (utilizing either
discrete or continuous features) by using additional visual information to limit comparisons. Concretely, a joint vocabulary is obtained from a combination of captioning systems. For a query word in this vocabulary, we perform alignment only between utterances whose images are captioned with that word, resulting in a set of extracted spoken segments.

The continuous-based system achieves the best overall results.
The discrete system achieves slightly worse performance, but is computationally more efficient.
Our best setup outperforms \herman{a neural baseline by roughly 23\%} %
in keyword localization %
and 31\% in keyword spotting.
\herman{We show qualitatively that failures occur because of co-occurring words which are impossible to disambiguate (\eg, ``box'' and ``ring'' from ``boxing ring'').
But even in idealized settings where
transcriptions are used, \dan{localization} performance}
\dan{is still not perfect,}
suggesting the %
\dan{need} for better ranking and alignment algorithms.

This work takes a key step toward bridging visual and spoken modalities for vocabulary discovery without text.
It shows promise for documenting low-resource or unwritten languages and supports efforts to model realistic cross-modal and cross-lingual associations \cite{azuh2019interspeech,oneata2024interspeech}.
In future work, we plan to apply these methods to
Hindi \cite{vaani2025,dharwarth2018_interlingua},
Yoruba \cite{olaleye2022yfaccyorubaspeechimagedataset}, or
Japanese \cite{ohishi2020icassp}
audio-image datasets.
\herman{In these cases, an English image captioner would still be used to filter the utterances, thereby uncovering English–foreign speech associations without transcriptions.}
Ultimately, our goal is to develop technology that supports the world's \textit{local} languages \cite{bird2022acl},
and, as such, we hope that this technology will help preserve and document the culture and heritage of these communities \cite{bird2024eacl, tapo2024interspeech}.

\bibliographystyle{IEEEbib}
\bibliography{refs}

\end{document}